\documentclass{article}

\PassOptionsToPackage{numbers, compress}{natbib}

\usepackage[preprint]{neurips_2024}

\usepackage[utf8]{inputenc} 
\usepackage[T1]{fontenc}    
\usepackage{hyperref}       
\usepackage{url}            
\usepackage{booktabs}       
\usepackage{amsfonts}       
\usepackage{nicefrac}       
\usepackage{microtype}      
\usepackage{xcolor}         

\usepackage{amsmath}
\usepackage{amssymb}
\usepackage{algorithm}
\usepackage{algpseudocode}
\usepackage{enumitem}
\usepackage{verbatim}
\usepackage{makecell}
\usepackage{pifont}
\usepackage{multirow}
\usepackage{tikz}
\usetikzlibrary{positioning}
\usetikzlibrary{arrows,shapes,chains}
\usepackage{graphicx}

\newcommand{\egno}{\textit{e.g.}}

\usepackage{marvosym}
\title{MambaCSR: Dual-Interleaved Scanning for Compressed Image Super-Resolution With SSMs}

\author{%
  Yulin Ren\textsuperscript{\rm 1}, Xin Li\textsuperscript{\rm 1\Letter}, Mengxi Guo\textsuperscript{\rm 2}, Bingchen Li\textsuperscript{\rm 1}, Shijie Zhao\textsuperscript{\rm 2} and Zhibo Chen\textsuperscript{\rm 1\Letter} \\
  \textsuperscript{\rm 1}University of Science and Technology of China, \textsuperscript{\rm 2}Bytedance Inc. \\
  \texttt{\small{\{renyulin, lbc31415926\}@mail.ustc.edu.cn}}, \\
\texttt{\small{\{xin.li, chenzhibo\}@ustc.edu.cn}}, \texttt{\small{\{guomengxi.qoelab, zhaoshijie.0526\}@bytedance.com}}
 }

\begin{document}

\maketitle
\renewcommand{\thefootnote}{}
\footnotetext{$^{\textrm{\Letter}}$  Xin Li and Zhibo Chen are corresponding authors.}
\begin{tikzpicture}[remember picture,overlay,shift={(current page.north west)}]
\node[anchor=north west,xshift=2.3cm,yshift=-2.4cm]{\scalebox{-1}[1]{\includegraphics[width=1.8cm]{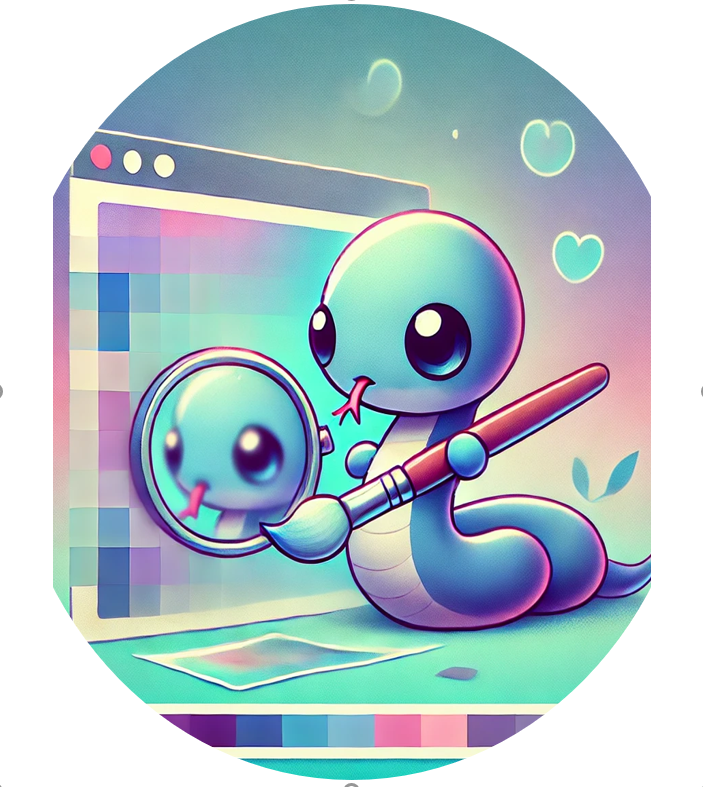}}};
\end{tikzpicture}
\begin{abstract}
We present MambaCSR, a simple but effective framework based on Mamba for the challenging compressed image super-resolution (CSR) task. Particularly, the scanning strategies of Mamba are crucial for effective contextual knowledge modeling in the restoration process despite it relying on selective state space modeling for all tokens. In this work, we propose an efficient dual-interleaved scanning paradigm (DIS) for CSR, which is composed of two scanning strategies: (i) hierarchical interleaved scanning is designed to comprehensively capture and utilize the most potential contextual information within an image by simultaneously taking advantage of the local window-based and sequential scanning methods; (ii) horizontal-to-vertical interleaved scanning is proposed to reduce the computational cost by leaving the redundancy between the scanning of different directions. To overcome the non-uniform compression artifacts, we also propose position-aligned cross-scale scanning to model multi-scale contextual information.   Experimental results on multiple benchmarks have shown the great performance of our MambaCSR in the compressed image super-resolution task. The code will be soon available in~\textcolor{magenta}{\url{https://github.com/renyulin-f/MambaCSR}}.

\end{abstract}

\section{Introduction}
Compressed Image Super-Resolution (CSR) has gradually emerged as an advanced task in industrial applications and human life, intending to remove the severe hybrid distortions caused by compression and low resolution simultaneously. In contrast to the existing single image super-resolution (SISR), CSR exhibits more non-uniform and diverse degradations including block artifacts, ringing effects, color drifting, etc., together with essential information loss. The above characteristics of CSR impose significant challenges for the contextual information modeling capability of existing super-resolution models.

A series of works have been explored to improve the contextual modeling capability through the framework design. The commonly used frameworks are usually based on three typical networks, including CNN~\cite{zhang2018image-RCAN,lim2017enhanced-EDSR,kim2016accurate-VDSR}, Transformer, and MLP. In particular, CNN excels at capturing local contextual information, while global contextual information must be aggregated by increasing the depth of networks. In contrast, Transformer-based works~\cite{conde2022swin2sr,li2022hst} utilize the self-attention mechanism to build long-range contextual dependencies for tokens across images, which entails a large computational cost. Unlike the above works, MLP-like works~\cite{li2024ucip} abandon the complex attention and successfully model the long-range contextual information with well-designed strategies for token mixers, significantly reducing the computational costs. Despite that, transformer-based works remain the mainstream for CSR tasks and deliver optimal performances. This raises a crucial question: ``whether there exists a new framework that can outperform transformers in CSR tasks".

\begin{figure}[t]
	\centering 
	\includegraphics[width=1\linewidth]{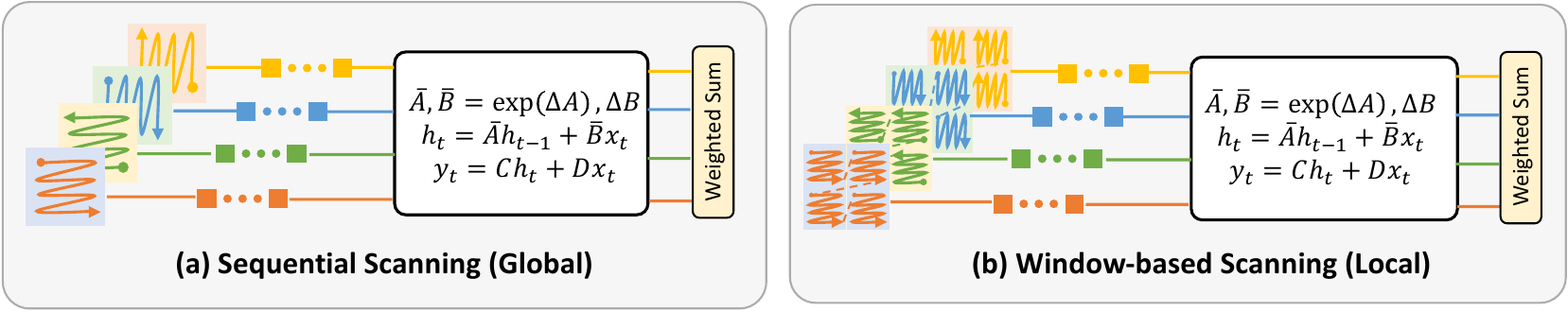}
	\caption{Comparisons of existing scanning methods. (a) Sequential scanning used in most low-level tasks~\cite{guo2024mambair,shi2024vmambair,shi2024multi-scale-mamba}. (b) Local window-based scanning in some high-level tasks~\cite{huang2024localmamba} }
	\label{fig1:scannning methods}
\end{figure}
Fortunately, Mamba, a brand-new framework~\cite{gu2023mamba} is proposed in the NLP field, which takes advantage of the selective State Space Model (SSM)~\cite{gu2021efficiently-S4}, thereby excelling at modeling long-range contextual information by dynamically deciding to preserve how much learned knowledge for each token in the scanning trajectory. Following this, lots of works have successfully applied the innovative framework to various visual fields, such as recognition~\cite{shen2024gamba-recognition}, segmentation~\cite{wan2024sigma-mamba-segmentation}, and generation~\cite{teng2024dim-generation}.  Thanks to the structure of SSM, the computational cost of Mamba is theoretically less than a transformer ($\mathcal{O}(n\log(n))$ versus $\mathcal{O}(n^2)$), which relieves the limitation of window-based representation learning in low-level vision with transformer. With the above advantages, some pioneering works have explored the Mamba framework for low-level vision tasks, \egno, image restoration~\cite{guo2024mambair,shi2024vmambair}, dehazing~\cite{zheng2024u-dehazing}, and deraining~\cite{zhen2024freqmamba-deraining}. However, the scanning strategies of the above works still follow early VMamba~\cite{liu2024vmamba} and rely on two horizontal and vertical scanning trajectories as shown in Fig.~\ref{fig1:scannning methods} (a) for long-range dependencies modeling, which tends to neglect the exploration of local dependencies.

However, in the context of the CSR task, the diverse and uni-formed hybrid degradations pose the high requirements of excavating the most informative contextual information across the whole tokens within one image. Consequently, both the local dependencies and long-range contextual information are crucial for CSR tasks, which motivated us to investigate how to design one scanning strategy to achieve the most comprehensive contextual modeling in Mamba.

In this work, we present MambaCSR, the first Mamba-based framework for CSR, intending to activate the comprehensive contextual modeling capability of Mamba through our proposed dual-interleaved scanning (DIS) strategy. Typically, window-based scanning shown in Fig.~\ref{fig1:scannning methods}(b) has been proven to be effective in capturing local dependencies for Mamba~\cite{huang2024localmamba,guan2024q-mamba}. 
Therefore, the hierarchical interleaved scanning of DIS is designed to iteratively apply window-based scanning and sequential-based scanning for MambaCSR, which aims to excavate both local and long-range contextual information simultaneously. From another perspective, the original VMamba exploits four types of scanning trajectories (\textit{i.e.}, two horizontal and vertical scanning strategies) for contextual modeling. However, not all scanning trajectories are necessary/essential for each token in each operation, thereby being redundant. To reduce the computational cost, we propose to decouple four scanning trajectories and iteratively exploit two horizontal and vertical scanning trajectories in the adjacent layer, resulting in the horizontal-to-vertical interleaved scanning for DIS. With our proposed dual-interleaved scanning paradigm, MambaCSR exhibits excellent contextual modeling capability and efficiency for CSR tasks.

To further overcome non-uniformed degradations in CSR, we also introduce a position-aligned cross-scale scanning strategy for CSR, aiming at fusing the multi-scale contextual information, thereby increasing the non-uniformed representation capability. Notably, a na\"ive method is scanning the features of the down-sampled image and its corresponding original image. However, it ignores that most relevant contextual information across different scales is usually distributed in the same region. This motivates us to scan the tokens in the same position across scales first and move the scanning windows of both scales together. The above scanning strategy further improves the restoration capability of MambaCSR for complicated degradations in CSR. The main contributions of this paper are summarized as follows:
\begin{itemize}
\item We present MambaCSR, the first Mamba-based framework for CSR tasks, which introduces the dual-interleaved scanning (DIS) paradigm, intending to activate more comprehensive and efficient contextual information modeling for MambaCSR. 
\item To achieve the DIS paradigm, we propose (i) hierarchical interleaved scanning to fuse the local and long-range contextual information and (ii) horizontal-to-vertical scanning to reduce the computational redundancy for the contextual modeling of different tokens. Moreover, we propose the position-aligned cross-scale scanning strategy to fuse the multi-scale contextual information, thereby eliminating the non-uniformed degradations of CSR. 
\item Experimental results on various compressed benchmarks demonstrate the effectiveness and efficiency of our proposed MambaCSR.
\end{itemize}

\section{Related Work}
\label{sec:related_work}

\subsection{Compressed Image Super-resolution}
Single image super-resolution (SISR) has been a long-standing research focus~\cite{chen2024ntire,wei2020aim-4,yue2024resshift,yang2024ntire-3,li2021learning-SR,wu2024seesr,qin2024quantsr,timofte2017ntire-panjinshan,wang2024exploiting-stablesr}. The pioneering work in this field is the CNN-based SRCNN~\cite{dong2015image-SRCNN}. Since then, numerous CNN-based models have emerged, incorporating deeper layers~\cite{kim2016accurate-VDSR,dong2016accelerating-FSRCNN,lim2017enhanced-EDSR,zhang2018learning-cnn}, residual blocks~\cite{zhang2018residual-RDN}, attention mechanisms~\cite{zhang2018image-RCAN,dai2019second-SAN,zhang2017learning-zuowang}, and non-local blocks~\cite{mei2021image-nonlocal,xia2022efficient-non-local2}, significantly advancing SR performance. However, CNNs still struggle to capture long-range contextual information dependencies. Following the introduction of self-attention, Transformer-based SR models~\cite{chen2021pre-ipt,lu2022transformer2,zhang2022accurate-transformer1,zhu2023attention-sr1,yu2023dipnet-sr2,yoo2023enriched-sr3} have been proposed, revolutionizing and dominating the SR field.

Based on this, compressed image super-resolution (CSR) has emerged as a new and crucial task closely tied to human life and industrial needs, originating from the AIM2022 competition~\cite{yang2022aim}. This task focuses on processing downsampled, compressed low-resolution images, which present more severe artifacts than SISR, thereby imposing higher demands on model capabilities~\cite{li2024promptcir,ren2024moe-diffir}. The pioneering work in the CSR field is Swin2SR~\cite{conde2022swin2sr}, based on the Swin-Transformer, which introduces several enhancements to SwinIR modules, improving training stability and better adapting to compressed distortions. Another notable example is HST~\cite{li2022hst}, employing a hierarchical Swin Transformer to fuse multi-scale compressed information. CIBDNet~\cite{qin2022cidbnet} introduces a dual-branch framework that combines convolutional and transformer branches, further boosting performance. UCIP~\cite{li2024ucip}, based on an MLP framework integrated with prompt learning, achieves universal CSR tasks with high efficiency. Recently, the state space model has shown great potential in long-range modeling with linear complexity, making it a promising candidate for CSR tasks.

\subsection{State Space Model in Vision Tasks}
Recently, the State Space Model has demonstrated high efficiency in capturing the dynamics and dependencies of long sequences. This has garnered significant attention in the visual field, covering various domains including 3D visual recognition~\cite{zhang2024point-3D,liang2024pointmamba-3D}, medical imaging~\cite{ma2024u-medical1,xie2024promamba}, multi-modal understanding~\cite{xu2024mambatalk-multimodal1,yang2024remamber-multimodal2}, remote sensing images~\cite{yao2024spectralmamba,he2024pan-remote1}, and more. Due to its impressive potential performance gains, SSM has also been introduced into low-level vision tasks~\cite{guo2024mambair,lei2024dvmsr-mamba2,zou2024wave-mamba3,wu2024rainmamba,fu2024ssumamba,di2024qmambabsr,dong2024mamba-Omamba,adhikarla2024expomamba,guo2024mambairv2attentivestatespace}. Among these, MambaIR~\cite{guo2024mambair} employs 2D Selective Scanning (SS2D)~\cite{liu2024vmamba} as the backbone while incorporating channel attention and convolutional layers to enhance local information capabilities. In contrast, VMambaIR~\cite{shi2024vmambair}, is based on a U-Net structure and extends channel dimension scanning for further feature extraction. FreqMamba~\cite{zhen2024freqmamba-deraining} operates from the frequency domain, utilizing Mamba blocks to achieve improved de-raining effects. LFMamba~\cite{lu2024lfmamba} employs an efficient SS2D mechanism to achieve light-field image-super-resolution. However, most existing Mamba approaches~\cite{cheng2024activating-mma} in the low-level field are preliminary explorations, relying primarily on the sequential scanning method. This approach overlooks the control over local information, which is critical in low-level tasks. In this paper, we introduce a dual-interleaved scanning method and a cross-scale scanning method to enhance the impact of local pixels and multi-scale features.

\begin{figure*}[t]
	\centering 
	\includegraphics[width=1\linewidth]{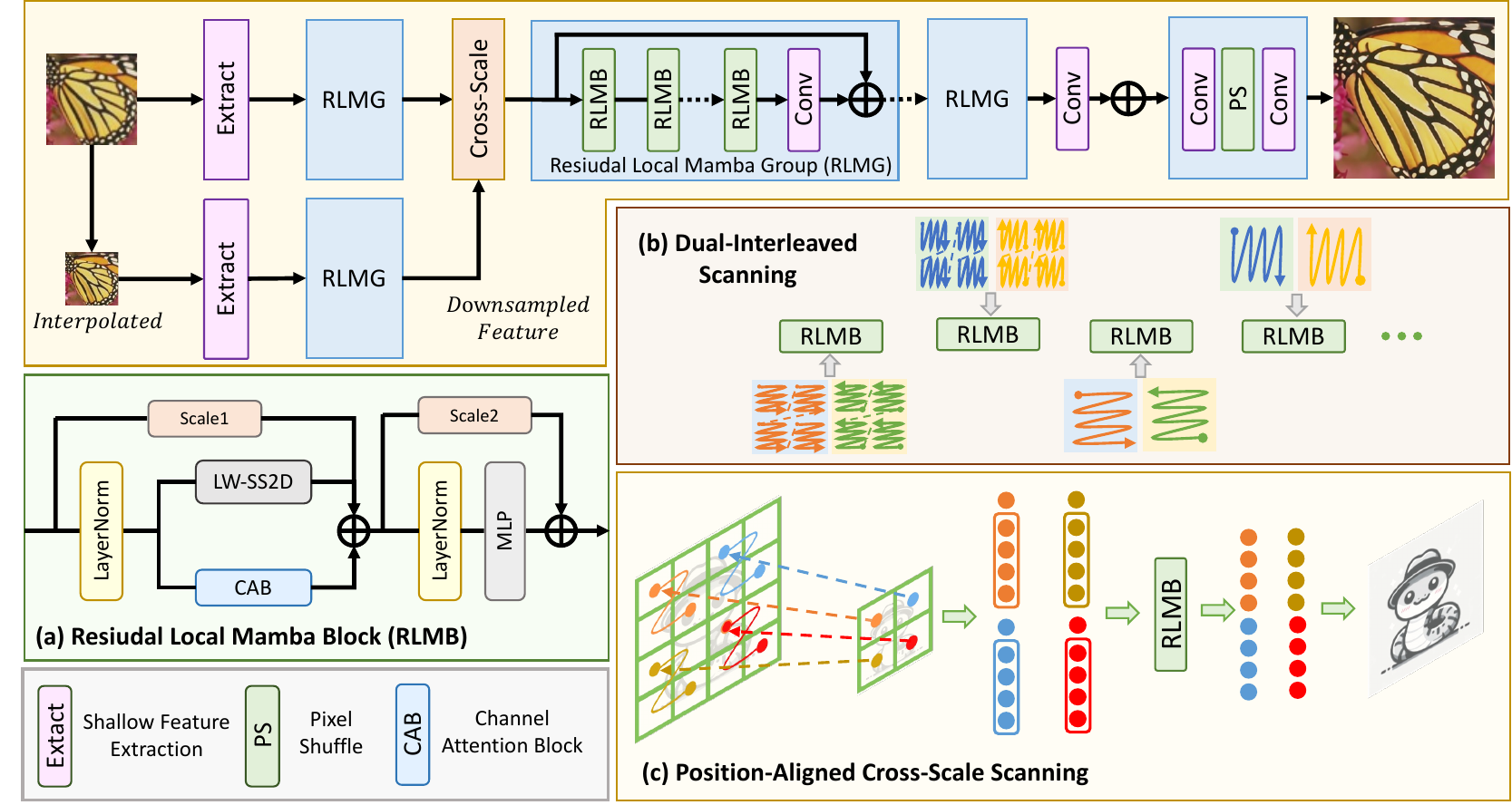}
	\caption{The Overall Architecture of MambaCSR. The overall pipeline (top) consists of three components: shallow feature extraction, deep feature extraction, and high-resolution reconstruction. (b) illustrates our proposed dual-interleaved scanning method. (c) represents the process of cross-scale scanning for fusing multi-scale features.}
	\label{fig:overall}
\end{figure*}

\section{Approaches}
\label{sec:Approaches}
In this section, we first review the basic theory of SSM in Sec.\ref{sec:preliminaries}. Then, we analyze the motivation of dual-interleaved scanning Sec.\ref{Sec: Motivation} from the perspective of contextual modeling within SSM. Next, we provide a detailed explanation of the dual-interleaved scanning and cross-scale scanning methods in Sec.\ref{sec:dual-interleaved scanning} and Sec.\ref{sec:Cross-scale scanning}, respectively. Finally, we introduce the overall framework of MambaCSR in Sec.~\ref{Sec: Network Architecture}.

\subsection{Preliminaries}
\label{sec:preliminaries}
\noindent\textbf{Preliminary.} State space model (SSM) is excellent at processing long linear sequences. Given a one-dimensional sequence input $x(t)$, SSM aims to output $y(t)$ through latent states $h(t)$ and input $x(t)$ as follows:
\begin{equation}
\begin{aligned}
    h'(t) &= Ah(t) + Bx(t), \\
    y(t) &= Ch(t) + Dx(t).
\end{aligned}
\label{equation_1}
\end{equation}
Here $A,B,C,D$ are four preset model parameters. As a discrete variant of SSM, mamba utilizes zero-order hold (ZOH) to discretize the process above while allowing an adaptive selective scanning mechanism for input data. The discretized values of $A$ and $B$ could be denoted as $\Bar{A}$ and $\Bar{B}$. Thus, Eq.~\ref{equation_1} could be reformulated as follows:
\begin{equation}
\begin{split}
\bar{A} &= e^{\Delta A}, \\
\bar{B} &= (\Delta A)^{-1}(e^{\Delta A} - I) \odot \Delta B, \\
h(t) &= \bar{A}h(t - 1) + \bar{B}x(t), \\
y(t) &= Ch(t) + Dx(t).
\end{split}
\end{equation}
Mamba aims to dynamically select $\Bar{A}$ and $\Bar{B}$ based on the input, incorporating parallel scanning to enhance computational efficiency.

\subsection{Motivations}
\label{Sec: Motivation}
Since the recurrent state space of Mamba, the contextual modeling in Mamba will vary for different scanning trajectories. To illustrate this, we can calculate the contribution of the $p$-th token on the $q$-th token following MSVMamba~\cite{shi2024multi-scale-mamba}:
\begin{equation}
\begin{aligned}
C_q \prod_{i=p}^{q} \bar{A}_i \bar{B}_p &= C_q \bar{A}_{(p \rightarrow q)} \bar{B}_p, \\
\bar{A}_{(p \rightarrow q)} &= e^{\sum_{i=p}^{q} \Delta_i \mathbf{A}}.
\end{aligned}
\label{equation: decay}
\end{equation}
The $\Delta_i \mathbf{A}$ in Eq.~\ref{equation: decay} is negative, indicating that as the number of scanned tokens increases, the contextual information of earlier scanned tokens on the current one is diminished. 
This also implies that the sequential scanning method shown in Fig.~\ref{fig1:scannning methods}(a) may overlook some local dependencies between adjacent tokens. In contrast, window-based scanning~\cite{huang2024localmamba} shown in Fig.~\ref{fig1:scannning methods}(b) excel at modeling local contextual information by narrowing the distance between adjacent tokens in the scanning trajectory with widow partition. 

A well-designed scanning strategy is crucial for the representation capability of the Mamba framework.  In terms of CSR tasks, the degradations are usually diverse, severe, and non-uniformed, which poses high requirements for comprehensive contextual information excavation.  To achieve this, we propose a hierarchical interleaved scanning approach that iteratively leverages the window-based and sequential scanning methods for joint excavation of local and long-range contextual information. Additionally, to eliminate the redundancy between four types of scanning directions for contextual modeling, we decouple it into interleaved horizontal and vertical scanning trajectories in adjacent layers, resulting in our horizontal-to-vertical scanning strategy. The above two scanning strategies contribute to our dual-interleaved scanning paradigm for CSR.

\subsection{Dual-Interleaved Scanning (DIS).}
\label{sec:dual-interleaved scanning}
As shown in Fig.~\ref{fig:overall}(b), we employ a dual-interleaved scanning approach to enhance the modeling of local dependencies and long-range contextual information, while also improving computational efficiency. This approach interleaves two types of scanning: (i) Hierarchical interleaved scanning is applied every two blocks to balance local and global feature extraction. (ii) Interleaved horizontal-to-vertical scanning. The original scanning method includes four scans, which introduces computational redundancy; consequently, one scanning path might dominate the critical information, while the others contribute less (The reason could be found in \textbf{Supplementary}). To address this, we reduce the four directional scans to two horizontal and two vertical scans, with each block undergoing only two scans. Crucially, we ensure that each scan is paired with its flipped counterpart. This pairing is necessary due to the non-causality of Mamba, where only preceding tokens contribute to subsequent tokens within a sequence. Paired scanning is therefore essential to ensure bidirectional information flow.

\subsection{Position-Aligned Cross-Scale Scanning.}
\label{sec:Cross-scale scanning} Compressed image super-resolution tasks necessitate the handling of non-uniform compression artifacts, which demand richer contextual information. In this paper, we propose a novel position-aligned cross-scale scanning method to deliver multi-scale contextual information, as illustrated in Fig.~\ref{fig:overall}(c). Specifically, as illustrated on the top-left side of Fig.~\ref{fig:overall}, the down-sampled image is generated through interpolation from the original image (scaling factor = 0.5), followed by shallow feature extraction and one RLMG layer. Each token in the down-sampled image corresponds to four tokens in the original image. Our core approach is to model four corresponding positions in the original feature map using each token in the down-sampled feature map. To clarify this process, we assume the down-sampled feature map is represented by the matrix:
\begin{equation}
\mathbf{X} = \begin{pmatrix}
\mathbf{x}^{(0,0)} & \mathbf{x}^{(0,1)} & \dots & \mathbf{x}^{(0,\mathcal{W}-1)} \\
\mathbf{x}^{(1,0)} & \mathbf{x}^{(1,1)} & \dots & \mathbf{x}^{(1,\mathcal{W}-1)} \\
\vdots & \vdots & \ddots & \vdots \\
\mathbf{x}^{(\mathcal{H}-1,0)} & \mathbf{x}^{(\mathcal{H}-1,1)} & \dots & \mathbf{x}^{(\mathcal{H}-1,\mathcal{W}-1)}
\end{pmatrix}
\end{equation}
The original feature map could be represented by:
\begin{equation}
\centering
\mathbf{Y} = \begin{pmatrix}
\mathbf{y}^{(0,0)} & \mathbf{y}^{(0,1)} & \cdots & \mathbf{y}^{(0,2\mathcal{W}-1)} \\
\mathbf{y}^{(1,0)} & \mathbf{y}^{(1,1)} & \cdots & \mathbf{y}^{(1,2\mathcal{W}-1)} \\
\vdots & \vdots & \ddots & \vdots \\
\mathbf{y}^{(2\mathcal{H}-1,0)} & \mathbf{y}^{(2\mathcal{H}-1,1)} & \cdots & \mathbf{y}^{(2\mathcal{H}-1,2\mathcal{W}-1)}
\end{pmatrix}
\end{equation}
For each token \( \mathbf{x}^{(i,j)} \) in the down-sampled feature, we sequentially scan the four corresponding positions in the original feature: \( \mathbf{Y}^{(2i,2j)} \), \( \mathbf{Y}^{(2i+1,2j)} \), \( \mathbf{Y}^{(2i,2j+1)} \), and \( \mathbf{Y}^{(2i+1,2j+1)} \) tokens. Then, we alternately scan the down-sampled features and the original features, unfold them into a 1D sequence, and feed this sequence into the RLMB for information interaction. This method ensures that the downsampled tokens, which exist at a 1:4 ratio within the 1D sequence, contribute effectively to the modeling of tokens in the original feature map. After processing through the Mamba Block, the tokens corresponding to the downsampled features are removed, retaining only those associated with the original image. This selective scanning approach optimizes the fusion of multi-scale features. Importantly, to conserve parameters and reduce computational complexity, this fusion process is applied only after the first RLMG block.

\subsection{Network Architecture}
\label{Sec: Network Architecture}
Followed by previous SR works~\cite{chen2023activating-HAT,liang2021swinir,lim2017enhanced-EDSR},  the overall architecture can be split into three components shown in the top of Fig.~\ref{fig:overall}: shallow feature extraction, deep feature extraction, and high-resolution reconstruction. Given a input compressed low resolution image $I_{LR} \in \mathbb{R}^{H \times W \times C_{in}}$, where $H$, $W$, $C_{in}$ denote the height, width, channel dimension of input $I_{LR}$. We first use a single convolution layer of kernel size $3 \times 3$ to extract shallow features of $I_{LR}$: 
\begin{equation}
    F_{0} = f_{sfe}(I_{LR})
\end{equation}
Then, we use several Residual Local Mamba Groups (RLMG) and one convolution layer to further extract deep features of $F_{0}$. The output deep features $F_{1} \in \mathbb{R}^{H \times W \times C}$ can be expressed as:
\begin{equation}
    F_{1} = f_{conv}(f_{rlmg}^{n}...(f_{rlmg}^{2}(f_{rlmg}^{1}(F_{0})))) + F_{0}
\end{equation}
We employ a residual connection to fuse the shallow feature $F_{0}$ with the deep feature $F_{1}$. Each RLMG consists of several Residual Local Mamba Blocks (RLMB), detailed in Sec.\ref{sec:RLMB}, along with a convolutional layer. Additionally, we generate a down-sampled image from the original using interpolation, process it through a convolutional layer and one RLMG, and then integrate it with the original image features via the position-aligned cross-scale module. Finally, for high-resolution reconstruction, we apply a pixel-shuffle layer to up-sample the fused feature. We optimize our MambaCSR following previous CSR works~\cite{li2022hst,conde2022swin2sr} with $L_{1}$ Loss:
\begin{equation}
    \mathcal{L} = \| I_{HR} - I_{LR} \|_1,
\end{equation}

\subsubsection*{Residual Local Mamba Block (RLMB).}
\label{sec:RLMB}
The Residual Local Mamba Block (RLMB) focuses on deep modeling of input features, capturing both local and long-range information. In each block, we first apply a normalization layer, followed by the adoption of a local-window-based 2D Selective Scan Module (LW-SS2D) to capture spatial long-range dependencies, as shown in Fig.~\ref{fig1:scannning methods} (a). We use different window sizes to represent local window-based and sequential scanning methods. For an input image size of $64 \times 64$ during the training phase, the window size for local scanning is set to 8, while the window size for sequential scanning is set to 64. Inspired by work HAT~\cite{chen2023activating-HAT}, we also employ channel attention block parallel to LW-SS2D to further enhance the representation ability of the network. Followed by MambaIR~\cite{guo2024mambair}, we use a scaling factor $s_{1}$ to control the intensity of input feature $X$ through skip connection.  After establishing long-term relationships, we apply Layer normalization and MLP to process the intermediate features. Scaling factor $s_{2}$ is also used to control the magnitude of residual. Thus, the entire process of RLMB could be expressed as:
\begin{equation}
\begin{aligned}
    F_{\text{inter}} &= \text{CAB}(\text{LN}(X)) + \text{LW-SSM}(\text{LN}(X)) + s_1 X \\
    F_{\text{out}} &= s_2 F_{\text{inter}} + \text{MLP}(\text{LN}(F_{\text{inter}}))
\end{aligned}
\end{equation}
Here, $F_{inter}$ denotes the intermediate features.

\section{Experiments}
\label{sec:Experiments}

\begin{table*}[t]
\centering
\caption{Quantitative comparison of compressed image super-resolution performance on JPEG~\cite{wallace1991jpeg} codec at QF levels [10, 20, 30] and HEVC~\cite{sze2014high-HEVC} codec at QP levels [32, 37, 42]. The highest performance values are highlighted in \textbf{bold}. Results are evaluated using PSNR$\uparrow$ and SSIM$\uparrow$ metrics. }
\resizebox{\textwidth}{!}{
\begin{tabular}{c|c|cc|cccccccccc}
\Xhline{2pt}
                               &                           & \multicolumn{2}{c|}{}                       & \multicolumn{10}{c}{Datasets}                                                                                                                                                                                                                                                                                                                                                                           \\ \cline{5-14} 
                               &                           & \multicolumn{2}{c|}{\multirow{-2}{*}{Cost}} & \multicolumn{2}{c|}{Set5}                                                        & \multicolumn{2}{c|}{Set14}                                                       & \multicolumn{2}{c|}{Manga109}                                                    & \multicolumn{2}{c|}{Urban100}                                                    & \multicolumn{2}{c}{DIV2K Test}                              \\ \cline{3-14} 
\multirow{-3}{*}{Codec} & \multirow{-3}{*}{Methods} & Params                & GFLOPS               & PSNR                         & \multicolumn{1}{c|}{SSIM}                         & PSNR                         & \multicolumn{1}{c|}{SSIM}                         & PSNR                         & \multicolumn{1}{c|}{SSIM}                         & PSNR                         & \multicolumn{1}{c|}{SSIM}                         & PSNR                         & SSIM                         \\ \hline
                                             & SwinIR                    & 8.72                  & 37.17               & {\color[HTML]{343434} 25.10} & \multicolumn{1}{c|}{{\color[HTML]{343434} 0.715}} & {\color[HTML]{343434} 23.97} & \multicolumn{1}{c|}{{\color[HTML]{343434} 0.603}} & {\color[HTML]{343434} 23.32} & \multicolumn{1}{c|}{{\color[HTML]{343434} 0.735}} & {\color[HTML]{343434} 22.03} & \multicolumn{1}{c|}{{\color[HTML]{343434} 0.600}} & {\color[HTML]{343434} 25.61} & {\color[HTML]{343434} 0.686} \\ \cline{2-14} 
                                             & Swin2SR                   & 11.67                 & 52.96               & {\color[HTML]{343434} 25.12} & \multicolumn{1}{c|}{{\color[HTML]{343434} 0.716}} & {\color[HTML]{343434} 23.96} & \multicolumn{1}{c|}{{\color[HTML]{343434} 0.603}} & {\color[HTML]{343434} 23.34} & \multicolumn{1}{c|}{{\color[HTML]{343434} 0.736}} & {\color[HTML]{343434} 22.05} & \multicolumn{1}{c|}{{\color[HTML]{343434} 0.601}} & {\color[HTML]{343434} 25.64} & {\color[HTML]{343434} 0.687} \\ \cline{2-14} 
                                             & HST                       & 16.48                 & 54.95               & {\color[HTML]{343434} 25.12} & \multicolumn{1}{c|}{{\color[HTML]{343434} 0.716}} & {\color[HTML]{343434} 24.00} & \multicolumn{1}{c|}{{\color[HTML]{343434} 0.603}} & {\color[HTML]{343434} 23.35} & \multicolumn{1}{c|}{{\color[HTML]{343434} 0.737}} & {\color[HTML]{343434} 22.09} & \multicolumn{1}{c|}{{\color[HTML]{343434} 0.602}} & {\color[HTML]{343434} 25.65} & {\color[HTML]{343434} 0.688} \\ \cline{2-14} 
                                             & HAT                       & 20.81                 & 102.4               & {\color[HTML]{343434} 25.13} & \multicolumn{1}{c|}{{\color[HTML]{343434} 0.718}} & {\color[HTML]{343434} 24.02} & \multicolumn{1}{c|}{{\color[HTML]{343434} 0.604}} & {\color[HTML]{343434} 23.42} & \multicolumn{1}{c|}{{\color[HTML]{343434} 0.784}} & {\color[HTML]{343434} 22.13} & \multicolumn{1}{c|}{{\color[HTML]{343434} 0.603}} & {\color[HTML]{343434} 25.66} & {\color[HTML]{343434} 0.690} \\ \cline{2-14} 
                                             & MambaIR                   & 16.70                 & 82.3                & {\color[HTML]{343434} 25.14} & \multicolumn{1}{c|}{{\color[HTML]{343434} 0.718}} & {\color[HTML]{343434} 24.01} & \multicolumn{1}{c|}{{\color[HTML]{343434} 0.605}} & {\color[HTML]{343434} 23.46} & \multicolumn{1}{c|}{{\color[HTML]{343434} 0.738}} & {\color[HTML]{343434} 22.14} & \multicolumn{1}{c|}{{\color[HTML]{343434} 0.604}} & {\color[HTML]{343434} 25.68} & {\color[HTML]{343434} 0.689} \\ \cline{2-14} 
\multirow{-6}{*}{JPEG (QF=10)}               & MambaCSR                  & 19.69                 & 76.03               & \textbf{25.20}               & \multicolumn{1}{c|}{\textbf{0.721}}               & \textbf{24.07}               & \multicolumn{1}{c|}{\textbf{0.607}}               & \textbf{23.57}               & \multicolumn{1}{c|}{\textbf{0.742}}               & \textbf{22.27}               & \multicolumn{1}{c|}{\textbf{0.608}}               & \textbf{25.74}               & \textbf{0.691}               \\ \hline
                                             & SwinIR                    & 8.72                  & 37.17               & {\color[HTML]{343434} 26.51} & \multicolumn{1}{c|}{{\color[HTML]{343434} 0.762}} & {\color[HTML]{343434} 25.03} & \multicolumn{1}{c|}{{\color[HTML]{343434} 0.639}} & {\color[HTML]{343434} 24.89} & \multicolumn{1}{c|}{{\color[HTML]{343434} 0.780}} & {\color[HTML]{343434} 23.04} & \multicolumn{1}{c|}{{\color[HTML]{343434} 0.646}} & {\color[HTML]{343434} 26.72} & {\color[HTML]{343434} 0.718} \\ \cline{2-14} 
                                             & Swin2SR                   & 11.67                 & 52.96               & {\color[HTML]{343434} 26.56} & \multicolumn{1}{c|}{{\color[HTML]{343434} 0.763}} & {\color[HTML]{343434} 25.05} & \multicolumn{1}{c|}{{\color[HTML]{343434} 0.639}} & {\color[HTML]{343434} 24.92} & \multicolumn{1}{c|}{{\color[HTML]{343434} 0.781}} & {\color[HTML]{343434} 23.07} & \multicolumn{1}{c|}{{\color[HTML]{343434} 0.647}} & {\color[HTML]{343434} 26.73} & {\color[HTML]{343434} 0.718} \\ \cline{2-14} 
                                             & HST                       & 16.48                 & 54.95               & {\color[HTML]{343434} 26.53} & \multicolumn{1}{c|}{{\color[HTML]{343434} 0.762}} & {\color[HTML]{343434} 25.02} & \multicolumn{1}{c|}{{\color[HTML]{343434} 0.639}} & {\color[HTML]{343434} 24.91} & \multicolumn{1}{c|}{{\color[HTML]{343434} 0.780}} & {\color[HTML]{343434} 23.06} & \multicolumn{1}{c|}{{\color[HTML]{343434} 0.646}} & {\color[HTML]{343434} 26.73} & {\color[HTML]{343434} 0.718} \\ \cline{2-14} 
                                             & HAT                       & 20.81                 & 102.4               & {\color[HTML]{343434} 26.49} & \multicolumn{1}{c|}{{\color[HTML]{343434} 0.762}} & {\color[HTML]{343434} 25.04} & \multicolumn{1}{c|}{{\color[HTML]{343434} 0.639}} & {\color[HTML]{343434} 24.98} & \multicolumn{1}{c|}{{\color[HTML]{343434} 0.784}} & {\color[HTML]{343434} 23.12} & \multicolumn{1}{c|}{{\color[HTML]{343434} 0.649}} & {\color[HTML]{343434} 26.74} & {\color[HTML]{343434} 0.719} \\ \cline{2-14} 
                                             & MambaIR                   & 16.70                 & 82.3                & {\color[HTML]{343434} 26.61} & \multicolumn{1}{c|}{{\color[HTML]{343434} 0.764}} & {\color[HTML]{343434} 25.11} & \multicolumn{1}{c|}{{\color[HTML]{343434} 0.640}} & {\color[HTML]{343434} 24.99} & \multicolumn{1}{c|}{{\color[HTML]{343434} 0.783}} & {\color[HTML]{343434} 23.13} & \multicolumn{1}{c|}{{\color[HTML]{343434} 0.653}} & {\color[HTML]{343434} 26.76} & {\color[HTML]{343434} 0.720} \\ \cline{2-14} 
\multirow{-6}{*}{JPEG (QF=20)}               & MambaCSR                  & 19.69                 & 76.03               & \textbf{26.67}               & \multicolumn{1}{c|}{\textbf{0.768}}               & \textbf{25.13}               & \multicolumn{1}{c|}{\textbf{0.642}}               & \textbf{25.12}               & \multicolumn{1}{c|}{\textbf{0.788}}               & \textbf{23.24}               & \multicolumn{1}{c|}{\textbf{0.658}}               & \textbf{26.79}               & \textbf{0.720}               \\ \hline
                                             & SwinIR                    & 8.72                  & 37.17               & {\color[HTML]{343434} 27.48} & \multicolumn{1}{c|}{{\color[HTML]{343434} 0.784}} & {\color[HTML]{343434} 25.61} & \multicolumn{1}{c|}{{\color[HTML]{343434} 0.659}} & {\color[HTML]{343434} 25.79} & \multicolumn{1}{c|}{{\color[HTML]{343434} 0.805}} & {\color[HTML]{343434} 23.59} & \multicolumn{1}{c|}{{\color[HTML]{343434} 0.672}} & {\color[HTML]{343434} 27.28} & {\color[HTML]{343434} 0.735} \\ \cline{2-14} 
                                             & Swin2SR                   & 11.67                 & 52.96               & {\color[HTML]{343434} 27.51} & \multicolumn{1}{c|}{{\color[HTML]{343434} 0.785}} & {\color[HTML]{343434} 25.61} & \multicolumn{1}{c|}{{\color[HTML]{343434} 0.659}} & {\color[HTML]{343434} 25.81} & \multicolumn{1}{c|}{{\color[HTML]{343434} 0.806}} & {\color[HTML]{343434} 23.63} & \multicolumn{1}{c|}{{\color[HTML]{343434} 0.673}} & {\color[HTML]{343434} 27.30} & {\color[HTML]{343434} 0.735} \\ \cline{2-14} 
                                             & HST                       & 16.48                 & 54.95               & {\color[HTML]{343434} 27.49} & \multicolumn{1}{c|}{{\color[HTML]{343434} 0.785}} & {\color[HTML]{343434} 25.60} & \multicolumn{1}{c|}{{\color[HTML]{343434} 0.659}} & {\color[HTML]{343434} 25.8}  & \multicolumn{1}{c|}{{\color[HTML]{343434} 0.805}} & {\color[HTML]{343434} 23.61} & \multicolumn{1}{c|}{{\color[HTML]{343434} 0.676}} & {\color[HTML]{343434} 27.31} & {\color[HTML]{343434} 0.736} \\ \cline{2-14} 
                                             & HAT                       & 20.81                 & 102.4               & {\color[HTML]{343434} 27.53} & \multicolumn{1}{c|}{{\color[HTML]{343434} 0.785}} & {\color[HTML]{343434} 25.61} & \multicolumn{1}{c|}{{\color[HTML]{343434} 0.660}} & {\color[HTML]{343434} 25.84} & \multicolumn{1}{c|}{{\color[HTML]{343434} 0.807}} & {\color[HTML]{343434} 23.67} & \multicolumn{1}{c|}{{\color[HTML]{343434} 0.677}} & {\color[HTML]{343434} 27.32} & {\color[HTML]{343434} 0.736} \\ \cline{2-14} 
                                             & MambaIR                   & 16.70                 & 82.3                & {\color[HTML]{343434} 27.56} & \multicolumn{1}{c|}{{\color[HTML]{343434} 0.786}} & {\color[HTML]{343434} 25.64} & \multicolumn{1}{c|}{{\color[HTML]{343434} 0.661}} & {\color[HTML]{343434} 25.87} & \multicolumn{1}{c|}{{\color[HTML]{343434} 0.808}} & {\color[HTML]{343434} 23.72} & \multicolumn{1}{c|}{{\color[HTML]{343434} 0.679}} & {\color[HTML]{343434} 27.34} & {\color[HTML]{343434} 0.737} \\ \cline{2-14} 
\multirow{-6}{*}{JPEG (QF=30)}               & MambaCSR                  & 19.69                 & 76.03               & \textbf{27.63}               & \multicolumn{1}{c|}{\textbf{0.788}}               & \textbf{25.70}               & \multicolumn{1}{c|}{\textbf{0.663}}               & \textbf{26.01}               & \multicolumn{1}{c|}{\textbf{0.812}}               & \textbf{23.82}               & \multicolumn{1}{c|}{\textbf{0.6825}}              & \textbf{27.39}               & \textbf{0.739}               \\ \hline
                                             & SwinIR                    & 8.72                  & 37.17               & {\color[HTML]{343434} 29.04} & \multicolumn{1}{c|}{{\color[HTML]{343434} 0.819}} & {\color[HTML]{343434} 26.71} & \multicolumn{1}{c|}{{\color[HTML]{343434} 0.705}} & {\color[HTML]{343434} 27.99} & \multicolumn{1}{c|}{{\color[HTML]{343434} 0.856}} & {\color[HTML]{343434} 24.76} & \multicolumn{1}{c|}{{\color[HTML]{343434} 0.723}} & {\color[HTML]{343434} 28.96} & {\color[HTML]{343434} 0.799} \\ \cline{2-14} 
                                             & Swin2SR                   & 11.67                 & 52.96               & {\color[HTML]{343434} 29.03} & \multicolumn{1}{c|}{{\color[HTML]{343434} 0.818}} & {\color[HTML]{343434} 26.68} & \multicolumn{1}{c|}{{\color[HTML]{343434} 0.703}} & {\color[HTML]{343434} 28.01} & \multicolumn{1}{c|}{{\color[HTML]{343434} 0.855}} & {\color[HTML]{343434} 24.79} & \multicolumn{1}{c|}{{\color[HTML]{343434} 0.725}} & {\color[HTML]{343434} 28.94} & {\color[HTML]{343434} 0.798} \\ \cline{2-14} 
                                             & HST                       & 16.48                 & 54.96               & {\color[HTML]{343434} 29.06} & \multicolumn{1}{c|}{{\color[HTML]{343434} 0.819}} & {\color[HTML]{343434} 26.71} & \multicolumn{1}{c|}{{\color[HTML]{343434} 0.703}} & {\color[HTML]{343434} 27.97} & \multicolumn{1}{c|}{{\color[HTML]{343434} 0.853}} & {\color[HTML]{343434} 24.82} & \multicolumn{1}{c|}{{\color[HTML]{343434} 0.726}} & {\color[HTML]{343434} 28.95} & {\color[HTML]{343434} 0.798} \\ \cline{2-14} 
                                             & HAT                       & 20.81                 & 102.4               & {\color[HTML]{343434} 29.11} & \multicolumn{1}{c|}{{\color[HTML]{343434} 0.821}} & {\color[HTML]{343434} 26.78} & \multicolumn{1}{c|}{{\color[HTML]{343434} 0.706}} & {\color[HTML]{343434} 28.15} & \multicolumn{1}{c|}{{\color[HTML]{343434} 0.861}} & {\color[HTML]{343434} 24.87} & \multicolumn{1}{c|}{{\color[HTML]{343434} 0.731}} & {\color[HTML]{343434} 28.98} & {\color[HTML]{343434} 0.801} \\ \cline{2-14} 
                                             & MambaIR                   & 16.70                 & 82.3                & {\color[HTML]{343434} 29.12} & \multicolumn{1}{c|}{{\color[HTML]{343434} 0.822}} & {\color[HTML]{343434} 26.81} & \multicolumn{1}{c|}{{\color[HTML]{343434} 0.707}} & {\color[HTML]{343434} 28.17} & \multicolumn{1}{c|}{{\color[HTML]{343434} 0.861}} & {\color[HTML]{343434} 24.89} & \multicolumn{1}{c|}{{\color[HTML]{343434} 0.732}} & {\color[HTML]{343434} 29.00} & {\color[HTML]{343434} 0.802} \\ \cline{2-14} 
\multirow{-6}{*}{HEVC (QP=32)}                 & MambaCSR                  & 19.69                 & 76.03               & \textbf{29.21}               & \multicolumn{1}{c|}{\textbf{0.824}}               & \textbf{26.86}               & \multicolumn{1}{c|}{\textbf{0.709}}               & \textbf{28.28}               & \multicolumn{1}{c|}{\textbf{0.867}}               & \textbf{25.02}               & \multicolumn{1}{c|}{\textbf{0.737}}               & \textbf{29.04}               & \textbf{0.803}               \\ \hline
                                             & SwinIR                    & 8.72                  & 37.17               & {\color[HTML]{343434} 27.12} & \multicolumn{1}{c|}{{\color[HTML]{343434} 0.774}} & {\color[HTML]{343434} 25.41} & \multicolumn{1}{c|}{{\color[HTML]{343434} 0.653}} & {\color[HTML]{343434} 23.55} & \multicolumn{1}{c|}{{\color[HTML]{343434} 0.673}} & {\color[HTML]{343434} 26.11} & \multicolumn{1}{c|}{{\color[HTML]{343434} 0.814}} & {\color[HTML]{343434} 26.84} & {\color[HTML]{343434} 0.730} \\ \cline{2-14} 
                                             & Swin2SR                   & 11.67                 & 52.96               & {\color[HTML]{343434} 27.13} & \multicolumn{1}{c|}{{\color[HTML]{343434} 0.773}} & {\color[HTML]{343434} 25.42} & \multicolumn{1}{c|}{{\color[HTML]{343434} 0.652}} & {\color[HTML]{343434} 23.58} & \multicolumn{1}{c|}{{\color[HTML]{343434} 0.672}} & {\color[HTML]{343434} 26.12} & \multicolumn{1}{c|}{{\color[HTML]{343434} 0.813}} & {\color[HTML]{343434} 26.83} & {\color[HTML]{343434} 0.728} \\ \cline{2-14} 
                                             & HST                       & 16.48                 & 54.96               & {\color[HTML]{343434} 27.09} & \multicolumn{1}{c|}{{\color[HTML]{343434} 0.773}} & {\color[HTML]{343434} 25.39} & \multicolumn{1}{c|}{{\color[HTML]{343434} 0.651}} & {\color[HTML]{343434} 23.54} & \multicolumn{1}{c|}{{\color[HTML]{343434} 0.671}} & {\color[HTML]{343434} 26.09} & \multicolumn{1}{c|}{{\color[HTML]{343434} 0.811}} & {\color[HTML]{343434} 26.83} & {\color[HTML]{343434} 0.728} \\ \cline{2-14} 
                                             & HAT                       & 20.81                 & 102.4               & {\color[HTML]{343434} 27.19} & \multicolumn{1}{c|}{{\color[HTML]{343434} 0.776}} & {\color[HTML]{343434} 25.46} & \multicolumn{1}{c|}{{\color[HTML]{343434} 0.655}} & {\color[HTML]{343434} 23.72} & \multicolumn{1}{c|}{{\color[HTML]{343434} 0.678}} & {\color[HTML]{343434} 26.21} & \multicolumn{1}{c|}{{\color[HTML]{343434} 0.819}} & {\color[HTML]{343434} 26.87} & {\color[HTML]{343434} 0.601} \\ \cline{2-14} 
                                             & MambaIR                   & 16.70                 & 82.3                & {\color[HTML]{343434} 27.18} & \multicolumn{1}{c|}{{\color[HTML]{343434} 0.777}} & {\color[HTML]{343434} 25.46} & \multicolumn{1}{c|}{{\color[HTML]{343434} 0.655}} & {\color[HTML]{343434} 23.74} & \multicolumn{1}{c|}{{\color[HTML]{343434} 0.680}} & {\color[HTML]{343434} 26.22} & \multicolumn{1}{c|}{{\color[HTML]{343434} 0.818}} & {\color[HTML]{343434} 26.89} & {\color[HTML]{343434} 0.732} \\ \cline{2-14} 
\multirow{-6}{*}{HEVC (QP=37)}                 & MambaCSR                  & 19.69                 & 76.03               & \textbf{27.22}               & \multicolumn{1}{c|}{\textbf{0.779}}               & \textbf{25.51}               & \multicolumn{1}{c|}{\textbf{0.658}}               & \textbf{23.83}               & \multicolumn{1}{c|}{\textbf{0.682}}               & \textbf{26.35}               & \multicolumn{1}{c|}{\textbf{0.822}}               & \textbf{26.94}               & \textbf{0.735}               \\ \hline
                                             & SwinIR                    & 8.72                  & 37.17               & {\color[HTML]{343434} 24.82} & \multicolumn{1}{c|}{{\color[HTML]{343434} 0.713}} & {\color[HTML]{343434} 23.80} & \multicolumn{1}{c|}{{\color[HTML]{343434} 0.598}} & {\color[HTML]{343434} 22.10} & \multicolumn{1}{c|}{{\color[HTML]{343434} 0.607}} & {\color[HTML]{343434} 23.81} & \multicolumn{1}{c|}{{\color[HTML]{343434} 0.752}} & {\color[HTML]{343434} 23.55} & {\color[HTML]{343434} 0.551} \\ \cline{2-14} 
                                             & Swin2SR                   & 11.67                 & 52.96               & {\color[HTML]{343434} 24.79} & \multicolumn{1}{c|}{{\color[HTML]{343434} 0.709}} & {\color[HTML]{343434} 23.78} & \multicolumn{1}{c|}{{\color[HTML]{343434} 0.597}} & {\color[HTML]{343434} 22.11} & \multicolumn{1}{c|}{{\color[HTML]{343434} 0.607}} & {\color[HTML]{343434} 23.79} & \multicolumn{1}{c|}{{\color[HTML]{343434} 0.753}} & {\color[HTML]{343434} 23.52} & {\color[HTML]{343434} 0.551} \\ \cline{2-14} 
                                             & HST                       & 16.48                 & 54.96               & {\color[HTML]{343434} 24.80} & \multicolumn{1}{c|}{{\color[HTML]{343434} 0.711}} & {\color[HTML]{343434} 23.77} & \multicolumn{1}{c|}{{\color[HTML]{343434} 0.597}} & {\color[HTML]{343434} 22.09} & \multicolumn{1}{c|}{{\color[HTML]{343434} 0.606}} & {\color[HTML]{343434} 23.78} & \multicolumn{1}{c|}{{\color[HTML]{343434} 0.750}} & {\color[HTML]{343434} 23.55} & {\color[HTML]{343434} 0.550} \\ \cline{2-14} 
                                             & HAT                       & 20.81                 & 102.4               & {\color[HTML]{343434} 24.86} & \multicolumn{1}{c|}{{\color[HTML]{343434} 0.714}} & {\color[HTML]{343434} 23.83} & \multicolumn{1}{c|}{{\color[HTML]{343434} 0.601}} & {\color[HTML]{343434} 22.24} & \multicolumn{1}{c|}{{\color[HTML]{343434} 0.614}} & {\color[HTML]{343434} 23.92} & \multicolumn{1}{c|}{{\color[HTML]{343434} 0.757}} & {\color[HTML]{343434} 23.59} & {\color[HTML]{343434} 0.553} \\ \cline{2-14} 
                                             & MambaIR                   & 16.70                 & 82.3                & {\color[HTML]{343434} 24.88} & \multicolumn{1}{c|}{{\color[HTML]{343434} 0.714}} & {\color[HTML]{343434} 23.83} & \multicolumn{1}{c|}{{\color[HTML]{343434} 0.731}} & {\color[HTML]{343434} 22.22} & \multicolumn{1}{c|}{{\color[HTML]{343434} 0.613}} & {\color[HTML]{343434} 23.94} & \multicolumn{1}{c|}{{\color[HTML]{343434} 0.758}} & {\color[HTML]{343434} 23.61} & {\color[HTML]{343434} 0.554} \\ \cline{2-14} 
\multirow{-6}{*}{HEVC (QP=42)}                 & MambaCSR                  & 19.69                 & 76.03               & \textbf{24.93}               & \multicolumn{1}{c|}{\textbf{0.719}}               & \textbf{23.87}               & \multicolumn{1}{c|}{\textbf{0.604}}               & \textbf{22.33}               & \multicolumn{1}{c|}{\textbf{0.618}}               & \textbf{24.03}               & \multicolumn{1}{c|}{\textbf{0.762}}               & \textbf{23.64}               & \textbf{0.554}        \\
\Xhline{2pt}
\end{tabular}
}
\label{tab: quantitative results}
\end{table*}

In this section, we first describe the dataset and implementation details in Sec.~\ref{Sec: dataset and implementation}. Next, we compare our proposed MambaCSR with current state-of-the-art CSR methods and present the visual results in Sec.~\ref{Sec: Comparison}. Finally, to validate the effectiveness of our proposed dual-interleaved scanning and cross-scale scanning method, we conduct a series of ablation studies, as detailed in Sec.~\ref{Sec: Ablation Study}.

\subsection{Dataset and Implementation Details}
\label{Sec: dataset and implementation}
We use the DF2K~\cite{Agustsson_2017_CVPR_Workshops-DIV2K} dataset as our primary training dataset. The DF2K training set is compressed using three widely used codecs: JPEG~\cite{wallace1991jpeg}, HEVC~\cite{sze2014high-HEVC}, and VVC~\cite{bross2021overview-VVC}. For each codec, we select three bitrate points as follows: (i) JPEG: quality factors (QF) of $[10, 20, 30]$, using the JPEG function provided by the OpenCV library; (ii) HEVC: quantization parameters (QP) of $[32, 37, 42]$, with compression conducted via the HM software; and (iii) VVC: QP values of $[32, 37, 42]$, using the VTM software for compression. For evaluation, we test on five commonly used super-resolution benchmarks: Set5~\cite{bevilacqua2012low-set5}, Set14~\cite{zeyde2012single-set14}, Manga109~\cite{matsui2017sketch-manga109}, Urban100~\cite{huang2015single-urban100}, and the DIV2K test set~\cite{Agustsson_2017_CVPR_Workshops-DIV2K}. The entire implementation is based on the PyTorch framework. During training, we employ the Adam optimizer ($\beta_{1}=0.9$, $\beta_{2}=0.99$) with an initial learning rate of $2e^{-4}$, which is decayed by a factor of 0.5 at $150k$, $225k$, and $275k$ iterations. Due to the small size of the DF2K dataset and its rapid convergence, the total number of iterations is set to $300k$. Input low-resolution images are randomly cropped to 64 $\times$ 64 and augmented with random flips and rotations. The window size for interleaved scanning is set to 8. The total batch size is set to 32, trained on eight NVIDIA V100 GPUs.

\subsection{Comparison with State-of-the-Art}
\label{Sec: Comparison}
We compare our proposed MambaCSR with two image super-resolution models (SwinIR~\cite{liang2021swinir} and HAT~\cite{chen2023activating-HAT}), two compressed image super-resolution models (Swin2SR~\cite{conde2022swin2sr} and HST~\cite{li2022hst}), and the Mamba-based baseline model, MambaIR~\cite{guo2024mambair}. All models are trained on the compressed DF2K dataset under the same settings. For evaluation, we use PSNR and SSIM metrics, with PSNR measured specifically on the Y channel.

\begin{figure*}[t]
	\centering 
	\includegraphics[width=1\linewidth]{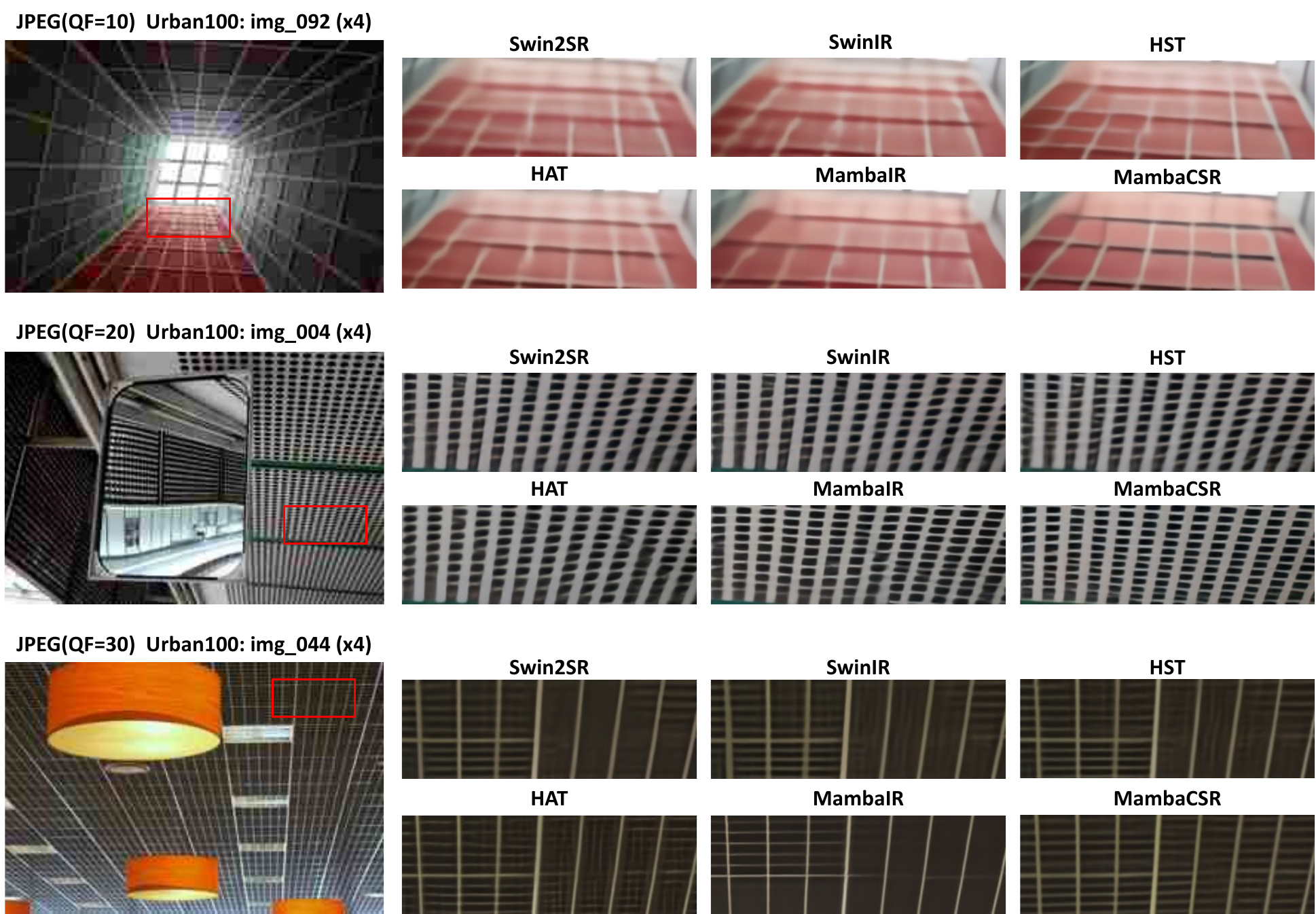}
	\caption{Visual comparison of our MambaCSR model with other state-of-the-art models on the Urban100~\cite{huang2015single-urban100} dataset under the JPEG codec at compression levels QF = [10, 20, 30]. Additional visual comparisons for HEVC and VVC codecs could be found in \textbf{Supplementary}.}
	\label{fig:visual_results}
\end{figure*}

\noindent \textbf{Quantitative Results.}
From Table~\ref{tab: quantitative results}, it is clear that our proposed MambaCSR achieves the highest performance across all five compressed benchmarks and at all compression levels for each codec in terms of PSNR and SSIM. Notably, on the Manga109 and Urban100 datasets, MambaCSR with cross-scale scanning surpasses MambaIR by an average of 0.12 dB, underscoring the effectiveness of our framework. This improvement is attributed to the introduction of dual-interleaved scanning, which enhances the model’s focus on local information, and cross-scale scanning, which better leverages multi-scale features to mitigate compression and downsampling artifacts. Additional results for VVC codec compression with QP values of [32, 37, 42] are provided in the \textbf{Supplementary}.

\noindent \textbf{Qualitative Results.}
We present visual comparisons with other models, showcasing three different compression levels (QF = [10, 20, 30]) under the JPEG codec. Additional visual comparisons for the HEVC and VVC codecs could be found in the \textbf{Supplementary}. As shown in Fig.~\ref{fig:visual_results}, our proposed MambaCSR demonstrates superior performance in handling compression artifacts, particularly in restoring textures and fine details, underscoring the robustness of our model.

\noindent \textbf{Computational Comparisons.}
As shown in Table~\ref{tab: quantitative results}, the computational cost of our proposed MambaCSR is 6 GFLOPS lower than that of MambaIR. Additionally, as shown in Table~\ref{new: ablation2}, when cross-scale scanning is omitted and only dual-interleaved scanning is applied, GFLOPS are reduced by approximately 11.

\begin{figure*}[t]
	\centering 
	\includegraphics[width=1\linewidth]{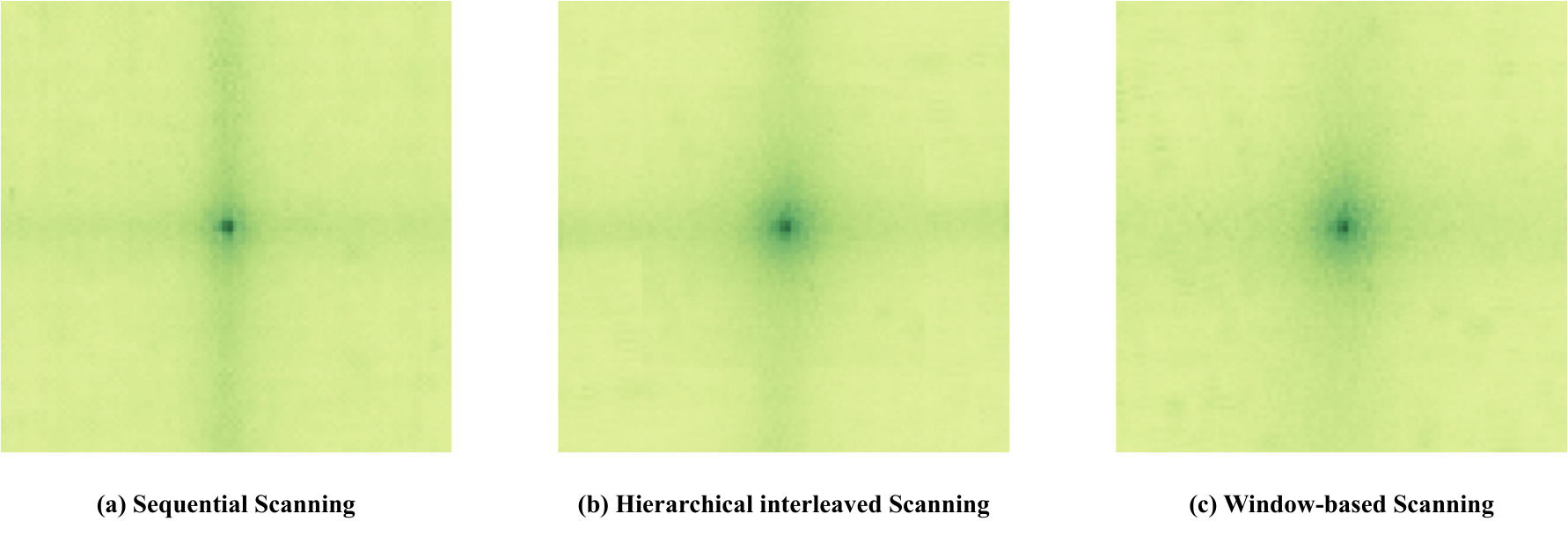}
	\caption{The receptive fields of different scanning methods: (a) Sequential scanning adopted in MambaIR, (b) Our hierarchical interleaved scanning, (c) Only local window-based scanning. All results are tested on the Manga109~\cite{matsui2017sketch-manga109} dataset. }
	\label{fig:ablation2}
\end{figure*}

\subsection{Ablation Studies}
\label{Sec: Ablation Study}

\begin{table}[t]
\centering
\caption{Effects of two cross-scale scanning methods: (I) our position-aligned cross-scale scanning. (II) simple sequential scanning. No scanning means only using dual-interleaved scanning.}
\label{new: ablation2}
\resizebox{\linewidth}{!}{ 
\begin{tabular}{c|c|cc|cc}
\Xhline{1.3pt}
\multirow{2}{*}{Scanning Methods} & \multirow{2}{*}{GFLOPs} & \multicolumn{2}{c|}{Manga109}          & \multicolumn{2}{c}{Urban100}          \\ \cline{3-6} 
 & & PSNR  & SSIM   & PSNR  & SSIM   \\ \hline
No scanning          & 70.45 & 23.53 & 0.7400 & 22.22 & 0.6071 \\
Position-Aligned     & 76.03 & \textbf{23.57} & \textbf{0.7419} & \textbf{22.26} & \textbf{0.6081} \\
Simple scanning      & 76.03 & 23.53 & 0.7403 & 22.24 & 0.6077 \\
\Xhline{1.3pt}
\end{tabular}
}
\end{table}

\noindent \textbf{Effects of different cross-scale scanning methods.}
In this paper, we introduce the position-aligned cross-scale scanning method, as shown in Fig.~\ref{fig:overall}(c). To validate the effectiveness of our position-aligned approach, we also implement a simple sequential scanning method, where all tokens of the low-resolution image are scanned first, followed by scanning the tokens of the high-resolution image. We then compare the performance of these two approaches. As shown in Table~\ref{new: ablation2}, the position-aligned cross-scale scanning method outperforms the sequential scanning approach. This indicates that the position-aligned method is more effective in modeling multi-scale features. The likely reason for the underperformance of the simple sequential scanning approach is that the long-sequence decay properties inherent to the SSM reduce the influence of low-resolution tokens on the original feature tokens.

\noindent \textbf{Effects of dual-interleaved scanning.} To validate the effectiveness of our dual-interleaved scanning method, we separately test the hierarchical interleaved scanning and horizontal-to-vertical interleaved scanning components within this framework. As shown in Table~\ref{new:ablation3}, hierarchical interleaved scanning yields the greatest performance gain, while horizontal-to-vertical scanning reduces 8 GFLOPs while maintaining comparable performance. The combination of these methods achieves a balance between performance and computational efficiency, demonstrating the dual-interleaved scanning method’s ability to enhance performance with reduced complexity. Additionally, we examine the receptive fields of different scanning methods, including sequential scanning, window-based scanning, and our proposed hierarchical-interleaved scanning. As illustrated in Fig.~\ref{fig:ablation2}, sequential scanning offers a smaller central receptive field while preserving some global information. Window-based scanning improves the central local receptive field but sacrifices global information modeling. In contrast, our hierarchical interleaved scanning effectively balances local and global receptive fields, retaining long-range modeling capacity while providing a larger local receptive field compared to sequential scanning.

\begin{table}[h!]
\centering
\caption{Effects of hierarchical interleaved scanning and horizontal-to-vertical interleaved scanning. The performance is tested on Manga109 and Urban100 datasets. The best performance is in \textbf{bold}.}
\resizebox{\linewidth}{!}{
\begin{tabular}{cc|c|cc}
\Xhline{2pt}
Hierarchical                                      & Horizontal-to-vertical                            & GFLOPs & Manga109 & Urban100 \\ \hline
{ \ding{55}} & { \ding{55}} & 84.56 & 23.44   & 22.12    \\
{ \ding{51}} & { \ding{55}} & 84.56 & 23.52   & 22.20    \\
{ \ding{55}} & { \ding{51}} & 76.03 & 23.43   & 22.12    \\
{ \ding{51}} & { \ding{51}} & 76.03 & \textbf{23.53}   & \textbf{22.22}    \\
\Xhline{2pt}
\end{tabular}
}
\label{new:ablation3}
\end{table}

\noindent \textbf{Effects of window-size used in dual-interleaved scanning.} We introduce an interleaved hierarchical scanning approach to capture both local and global information from compressed images. In window-based local scanning, images are divided into regions according to the window size. To determine the optimal window size, we conduct ablation studies without cross-scale scanning. Given that the input low-resolution images during training are 64 $\times$ 64,  we test five window sizes: 64 $\times$ 64, 32 $\times$ 32, 16 $\times$ 16, 8 $\times$ 8, and 4 $\times$ 4, and tested the results on Manga109 and Urban100 datasets. As shown in Table~\ref{new: ablation1}, the 8 $\times$ 8 window size strikes the best balance for effectively capturing local information.

\begin{table}[t]
\centering
\caption{The effect of window size used in dual-interleaved scanning. The best performance is in \textbf{bold}.}
\label{new: ablation1}
\begin{tabular}{c|cc|cc}
\Xhline{1.3pt}
                              & \multicolumn{2}{c|}{Manga109}                                & \multicolumn{2}{c}{Urban100}                                 \\ \cline{2-5} 
\multirow{-2}{*}{Window Size} & PSNR                         & SSIM                          & PSNR                         & SSIM                          \\ \hline
4 $\times$ 4                           & { 23.49} & { 0.7394} & { 22.19} & { 0.6059} \\
8 $\times$ 8                           & { \textbf{23.53}} & { \textbf{0.7400}} & { \textbf{22.22}} & { \textbf{0.6071}} \\
16 $\times$ 16                         & { 23.51} & { 0.7396} & { 22.20} & { 0.6065} \\
32 $\times$ 32                         & { 23.47} & { 0.7377} & { 22.17} & { 0.6043} \\
64 $\times$ 64                         & { 23.47}                        & { 0.7374}                        & { 22.14}                        & { 0.6036}   \\
\Xhline{1.3pt}
\end{tabular}
\end{table}

\section{Conclusion}
In this paper, we introduce MambaCSR, the first Mamba-based framework for compressed image super-resolution. Concretely, we investigate the effects of different scanning strategies for the contextual modeling of CSR, and propose the efficient dual-interleaved scanning (DIS) paradigm for CSR, intending to activate comprehensive contextual modeling of Mamba. The DIS is composed of hierarchical interleaved scanning, which aims to excavate the local and long-range contextual information simultaneously, and horizontal-to-vertical interleaved scanning to reduce the computational redundancy caused by four types of scanning trajectories in Mamba. To overcome the non-uniformed compression degradation and further enhance CSR performance, we develop a novel position-aligned cross-scale scanning method that effectively fuses multi-scale features. The experimental results across various compressed benchmarks demonstrate the advantages of our dual-interleaved scanning approach and validate the superiority of the position-aligned cross-scale scanning method. We also analyze the limitations and broad impacts of our MambaCSR in the \textbf{supplementary}.



\bibliography{main}
\bibliographystyle{abbrv}


\clearpage
\newpage
\appendix

\section{Supplementary Material}
In this document, we first describe the broader impact and limitations of our paper in Sec.~\ref{broder impact} and Sec.~\ref{limitations} respectively. After that, we provide more quantitative results in Sec.~\ref{quantitative results}. Finally, we show more visual comparisons in Sec.~\ref{visual results}.

\subsection{Broader Impact}
\label{broder impact}
We propose MambaCSR in this paper, which introduces an innovative dual-interleaved scanning paradigm for compressed image super-resolution (CSR). By leveraging hierarchical interleaved scanning and horizontal-to-vertical scanning, MambaCSR significantly reduces redundancy while enhancing contextual information modeling, leading to more efficient and accurate super-resolution. These advancements not only improve the performance of CSR tasks in industrial applications but also contribute to reducing the computational burden associated with processing high-resolution images. The primary applications of MambaCSR span a range of areas, including image restoration, multimedia content enhancement, and visual communication. Importantly, the deployment of MambaCSR adheres to ethical standards, ensuring that the technology is used responsibly to avoid potential biases and harms in visual data processing. By focusing on efficiency and accuracy, MambaCSR paves the way for more sustainable and equitable advancements in the field of image processing.

\subsection{Limitations}
\label{limitations}
In this work, we introduce MambaCSR and propose dual-interleaved, position-aligned cross-scale scanning methods. While our approach demonstrates state-of-the-art performance on JPEG, HEVC and VVC compressed image super-resolution tasks, the generalization of MambaCSR to other compression standards remains unexplored. The CSR field encompasses a wide range of compression methods beyond JPEG, VVC~\cite{bross2021overview-VVC} and HEVC~\cite{sze2014high-HEVC} including traditional codecs such as WebP~\cite{ginesu2012objective-webp} and AVC~\cite{wiegand2003overview-AVC}, as well as newer, learning-based codecs~\cite{wu2021learned-code2,hu2020coarse-wenhan-yang} like HIFIC~\cite{mentzer2020high-HIFIC}.  The applicability of MambaCSR across these diverse compression techniques is still an open question. Future work will aim to evaluate MambaCSR’s performance on various compression standards and explore universal restoration methods. Additionally, the current training speed of MambaCSR for low-level tasks is relatively slow. Addressing this limitation will involve investigating more efficient training strategies and network designs to enhance MambaCSR’s training efficiency in future research.

\begin{table*}[h]
\centering
\caption{Quantitative comparison of compressed image super-resolution performance on JPEG~\cite{wallace1991jpeg} codec at QF levels [10, 20, 30] and HEVC~\cite{sze2014high-HEVC} codec at QP levels [32, 37, 42]. The highest performance values are highlighted in \textbf{bold}. Results are evaluated using PSNR$\uparrow$ and SSIM$\uparrow$ metrics. }
\resizebox{\textwidth}{!}{
\begin{tabular}{c|c|cc|cccccccccc}
\Xhline{2pt}
                              &                           & \multicolumn{2}{c|}{}                       & \multicolumn{10}{c}{Datasets}                                                                                                                                                                                                                                                                                                                                                                           \\ \cline{5-14} 
                              &                           & \multicolumn{2}{c|}{\multirow{-2}{*}{Cost}} & \multicolumn{2}{c|}{Set5}                                                        & \multicolumn{2}{c|}{Set14}                                                       & \multicolumn{2}{c|}{Manga109}                                                    & \multicolumn{2}{c|}{Urban100}                                                    & \multicolumn{2}{c}{DIV2K Test}                              \\ \cline{3-14} 
\multirow{-3}{*}{Codec}       & \multirow{-3}{*}{Methods} & Params                & Flops               & PSNR                         & \multicolumn{1}{c|}{SSIM}                         & PSNR                         & \multicolumn{1}{c|}{SSIM}                         & PSNR                         & \multicolumn{1}{c|}{SSIM}                         & PSNR                         & \multicolumn{1}{c|}{SSIM}                         & PSNR                         & SSIM                         \\ \hline
                              & SwinIR~\cite{liang2021swinir}                   & 8.72                  & 37.17               & {\color[HTML]{343434} 29.06} & \multicolumn{1}{c|}{{\color[HTML]{343434} 0.822}} & {\color[HTML]{343434} 26.84} & \multicolumn{1}{c|}{{\color[HTML]{343434} 0.709}} & {\color[HTML]{343434} 28.12} & \multicolumn{1}{c|}{{\color[HTML]{343434} 0.860}} & {\color[HTML]{343434} 24.83} & \multicolumn{1}{c|}{{\color[HTML]{343434} 0.730}} & {\color[HTML]{343434} 28.64} & {\color[HTML]{343434} 0.795} \\ \cline{2-14} 
                              & Swin2SR~\cite{conde2022swin2sr}                  & 11.67                 & 52.96               & {\color[HTML]{343434} 29.04} & \multicolumn{1}{c|}{{\color[HTML]{343434} 0.822}} & {\color[HTML]{343434} 26.82} & \multicolumn{1}{c|}{{\color[HTML]{343434} 0.706}} & {\color[HTML]{343434} 28.13} & \multicolumn{1}{c|}{{\color[HTML]{343434} 0.858}} & {\color[HTML]{343434} 24.81} & \multicolumn{1}{c|}{{\color[HTML]{343434} 0.727}} & {\color[HTML]{343434} 28.63} & {\color[HTML]{343434} 0.793} \\ \cline{2-14} 
                              & HST~\cite{li2022hst}                      & 16.48                 & 54.96               & {\color[HTML]{343434} 29.07} & \multicolumn{1}{c|}{{\color[HTML]{343434} 0.821}} & {\color[HTML]{343434} 26.79} & \multicolumn{1}{c|}{{\color[HTML]{343434} 0.708}} & {\color[HTML]{343434} 28.11} & \multicolumn{1}{c|}{{\color[HTML]{343434} 0.857}} & {\color[HTML]{343434} 24.79} & \multicolumn{1}{c|}{{\color[HTML]{343434} 0.726}} & {\color[HTML]{343434} 28.65} & {\color[HTML]{343434} 0.795} \\ \cline{2-14} 
                              &HAT~\cite{chen2023activating-HAT}                       & 20.81                 & 102.4               & {\color[HTML]{343434} 29.09} & \multicolumn{1}{c|}{{\color[HTML]{343434} 0.824}} & {\color[HTML]{343434} 26.87} & \multicolumn{1}{c|}{{\color[HTML]{343434} 0.711}} & {\color[HTML]{343434} 28.19} & \multicolumn{1}{c|}{{\color[HTML]{343434} 0.862}} & {\color[HTML]{343434} 24.97} & \multicolumn{1}{c|}{{\color[HTML]{343434} 0.737}} & {\color[HTML]{343434} 28.67} & {\color[HTML]{343434} 0.796} \\ \cline{2-14} 
                              & MambaIR~\cite{guo2024mambair}                  & 16.70                 & 82.3                & {\color[HTML]{343434} 29.11} & \multicolumn{1}{c|}{{\color[HTML]{343434} 0.823}} & {\color[HTML]{343434} 26.87} & \multicolumn{1}{c|}{{\color[HTML]{343434} 0.710}} & {\color[HTML]{343434} 28.22} & \multicolumn{1}{c|}{{\color[HTML]{343434} 0.861}} & {\color[HTML]{343434} 24.96} & \multicolumn{1}{c|}{{\color[HTML]{343434} 0.734}} & {\color[HTML]{343434} 28.68} & {\color[HTML]{343434} 0.797} \\ \cline{2-14} 
\multirow{-6}{*}{VVC (QP=32)} & MambaCSR                  & 19.69                 & 76.03               & \textbf{29.15}               & \multicolumn{1}{c|}{\textbf{0.827}}               & \textbf{26.93}               & \multicolumn{1}{c|}{\textbf{0.714}}               & \textbf{28.31}               & \multicolumn{1}{c|}{\textbf{0.864}}               & \textbf{25.08}               & \multicolumn{1}{c|}{\textbf{0.741}}               & \textbf{28.73}               & \textbf{0.799}               \\ \hline
                              & SwinIR~\cite{liang2021swinir}                   & 8.72                  & 37.17               & {\color[HTML]{343434} 27.21} & \multicolumn{1}{c|}{{\color[HTML]{343434} 0.776}} & {\color[HTML]{343434} 23.53} & \multicolumn{1}{c|}{{\color[HTML]{343434} 0.658}} & {\color[HTML]{343434} 26.27} & \multicolumn{1}{c|}{{\color[HTML]{343434} 0.818}} & {\color[HTML]{343434} 23.68} & \multicolumn{1}{c|}{{\color[HTML]{343434} 0.679}} & {\color[HTML]{343434} 26.99} & {\color[HTML]{343434} 0.756} \\ \cline{2-14} 
                              & Swin2SR~\cite{conde2022swin2sr}                  & 11.67                 & 52.96               & {\color[HTML]{343434} 27.23} & \multicolumn{1}{c|}{{\color[HTML]{343434} 0.777}} & {\color[HTML]{343434} 25.51} & \multicolumn{1}{c|}{{\color[HTML]{343434} 0.654}} & {\color[HTML]{343434} 26.25} & \multicolumn{1}{c|}{{\color[HTML]{343434} 0.818}} & {\color[HTML]{343434} 23.67} & \multicolumn{1}{c|}{{\color[HTML]{343434} 0.679}} & {\color[HTML]{343434} 26.96} & {\color[HTML]{343434} 0.756} \\ \cline{2-14} 
                              & HST~\cite{li2022hst}                      & 16.48                 & 54.96               & {\color[HTML]{343434} 27.20} & \multicolumn{1}{c|}{{\color[HTML]{343434} 0.776}} & {\color[HTML]{343434} 25.52} & \multicolumn{1}{c|}{{\color[HTML]{343434} 0.657}} & {\color[HTML]{343434} 26.25} & \multicolumn{1}{c|}{{\color[HTML]{343434} 0.817}} & {\color[HTML]{343434} 23.67} & \multicolumn{1}{c|}{{\color[HTML]{343434} 0.678}} & {\color[HTML]{343434} 27.00} & {\color[HTML]{343434} 0.755} \\ \cline{2-14} 
                              &HAT~\cite{chen2023activating-HAT}                       & 20.81                 & 102.4               & {\color[HTML]{343434} 27.24} & \multicolumn{1}{c|}{{\color[HTML]{343434} 0.779}} & {\color[HTML]{343434} 25.55} & \multicolumn{1}{c|}{{\color[HTML]{343434} 0.658}} & {\color[HTML]{343434} 26.30} & \multicolumn{1}{c|}{{\color[HTML]{343434} 0.819}} & {\color[HTML]{343434} 23.74} & \multicolumn{1}{c|}{{\color[HTML]{343434} 0.682}} & {\color[HTML]{343434} 27.01} & {\color[HTML]{343434} 0.757} \\ \cline{2-14} 
                              & MambaIR~\cite{guo2024mambair}                  & 16.70                 & 82.3                & {\color[HTML]{343434} 27.26} & \multicolumn{1}{c|}{{\color[HTML]{343434} 0.779}} & {\color[HTML]{343434} 25.57} & \multicolumn{1}{c|}{{\color[HTML]{343434} 0.659}} & {\color[HTML]{343434} 26.32} & \multicolumn{1}{c|}{{\color[HTML]{343434} 0.821}} & {\color[HTML]{343434} 23.77} & \multicolumn{1}{c|}{{\color[HTML]{343434} 0.684}} & {\color[HTML]{343434} 27.03} & {\color[HTML]{343434} 0.758} \\ \cline{2-14} 
\multirow{-6}{*}{VVC (QP=37)} & MambaCSR                  & 19.69                 & 76.03               & \textbf{27.30}               & \multicolumn{1}{c|}{\textbf{0.783}}               & \textbf{25.62}               & \multicolumn{1}{c|}{\textbf{0.661}}               & \textbf{26.43}               & \multicolumn{1}{c|}{\textbf{0.824}}               & \textbf{23.85}               & \multicolumn{1}{c|}{\textbf{0.686}}               & \textbf{27.06}               & \textbf{0.761}               \\ \hline
                              & SwinIR~\cite{liang2021swinir}                   & 8.72                  & 37.17               & {\color[HTML]{343434} 25.12} & \multicolumn{1}{c|}{{\color[HTML]{343434} 0.721}} & {\color[HTML]{343434} 23.93} & \multicolumn{1}{c|}{{\color[HTML]{343434} 0.603}} & {\color[HTML]{343434} 24.08} & \multicolumn{1}{c|}{{\color[HTML]{343434} 0.762}} & {\color[HTML]{343434} 22.27} & \multicolumn{1}{c|}{{\color[HTML]{343434} 0.615}} & {\color[HTML]{343434} 24.73} & {\color[HTML]{343434} 0.657} \\ \cline{2-14} 
                              & Swin2SR~\cite{conde2022swin2sr}                  & 11.67                 & 52.96               & {\color[HTML]{343434} 25.11} & \multicolumn{1}{c|}{{\color[HTML]{343434} 0.720}} & {\color[HTML]{343434} 23.91} & \multicolumn{1}{c|}{{\color[HTML]{343434} 0.602}} & {\color[HTML]{343434} 24.06} & \multicolumn{1}{c|}{{\color[HTML]{343434} 0.761}} & {\color[HTML]{343434} 22.25} & \multicolumn{1}{c|}{{\color[HTML]{343434} 0.612}} & {\color[HTML]{343434} 24.72} & {\color[HTML]{343434} 0.655} \\ \cline{2-14} 
                              & HST~\cite{li2022hst}                      & 16.48                 & 54.96               & {\color[HTML]{343434} 25.14} & \multicolumn{1}{c|}{{\color[HTML]{343434} 0.722}} & {\color[HTML]{343434} 23.95} & \multicolumn{1}{c|}{{\color[HTML]{343434} 0.603}} & {\color[HTML]{343434} 24.08} & \multicolumn{1}{c|}{{\color[HTML]{343434} 0.760}} & {\color[HTML]{343434} 22.28} & \multicolumn{1}{c|}{{\color[HTML]{343434} 0.616}} & {\color[HTML]{343434} 24.74} & {\color[HTML]{343434} 0.657} \\ \cline{2-14} 
                              &HAT~\cite{chen2023activating-HAT}                       & 20.81                 & 102.4               & {\color[HTML]{343434} 25.14} & \multicolumn{1}{c|}{{\color[HTML]{343434} 0.722}} & {\color[HTML]{343434} 23.95} & \multicolumn{1}{c|}{{\color[HTML]{343434} 0.604}} & {\color[HTML]{343434} 24.12} & \multicolumn{1}{c|}{{\color[HTML]{343434} 0.763}} & {\color[HTML]{343434} 22.31} & \multicolumn{1}{c|}{{\color[HTML]{343434} 0.619}} & {\color[HTML]{343434} 24.74} & {\color[HTML]{343434} 0.658} \\ \cline{2-14} 
                              & MambaIR~\cite{guo2024mambair}                  & 16.70                 & 82.3                & {\color[HTML]{343434} 25.16} & \multicolumn{1}{c|}{{\color[HTML]{343434} 0.724}} & {\color[HTML]{343434} 23.99} & \multicolumn{1}{c|}{{\color[HTML]{343434} 0.605}} & {\color[HTML]{343434} 24.20} & \multicolumn{1}{c|}{{\color[HTML]{343434} 0.765}} & {\color[HTML]{343434} 22.25} & \multicolumn{1}{c|}{{\color[HTML]{343434} 0.622}} & {\color[HTML]{343434} 24.76} & {\color[HTML]{343434} 0.659} \\ \cline{2-14} 
\multirow{-6}{*}{VVC (QP=42)} & MambaCSR                  & 19.69                 & 76.03               & \textbf{25.18}               & \multicolumn{1}{c|}{\textbf{0.725}}               & \textbf{24.03}               & \multicolumn{1}{c|}{\textbf{0.606}}               & \textbf{24.26}               & \multicolumn{1}{c|}{\textbf{0.767}}               & \textbf{22.42}               & \multicolumn{1}{c|}{\textbf{0.621}}               & \textbf{24.77}               & \textbf{0.661}    \\
\Xhline{2pt}
\end{tabular}
}
\label{tab: VVC_Quanti}
\end{table*}

\subsection{More Quantitative Results}
\label{quantitative results}
In this section, we present additional quantitative results that complement the main text, focusing specifically on the VVC codec under QP settings of [32, 37, 42]. Our approach is compared against four transformer-based models: SwinIR~\cite{liang2021swinir}, Swin2SR~\cite{conde2022swin2sr}, HST~\cite{li2022hst}, and HAT~\cite{chen2023activating-HAT}, as well as the Mamba-based model, MambaIR~\cite{guo2024mambair}. We adopt DF2K as our primary training dataset~\cite{Agustsson_2017_CVPR_Workshops-DIV2K}. The evaluation is performed on five compressed benchmarks: Set5~\cite{bevilacqua2012low-set5}, Set14~\cite{zeyde2012single-set14}, Manga109~\cite{matsui2017sketch-manga109}, Urban100~\cite{huang2015single-urban100}, and the DIV2K test set~\cite{Agustsson_2017_CVPR_Workshops-DIV2K}, with all datasets compressed using VTM software. 

As shown in Table~\ref{tab: VVC_Quanti}, our model achieves state-of-the-art performance across all benchmarks, significantly surpassing MambaIR by an average of approximately 0.11–0.12 dB on the Manga109 and Urban100 datasets. Moreover, our approach reduces computational complexity by 6 GFLOPS compared to MambaIR, underscoring both the efficiency and effectiveness of our proposed method.

\subsection{More Visual Comparisons}
\label{visual results}
\begin{figure*}[h]
	\centering 
	\includegraphics[width=1\linewidth]{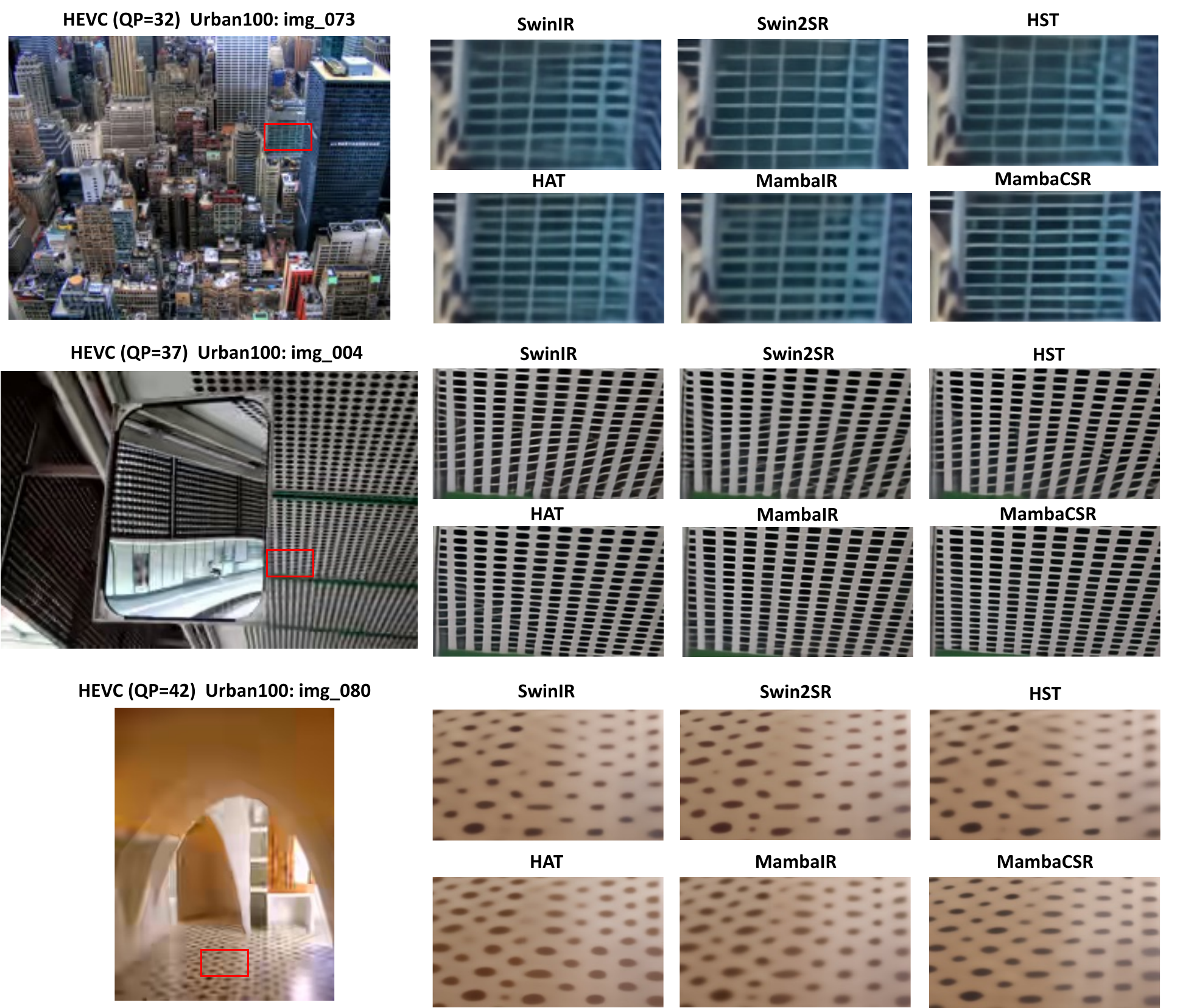}
	\caption{Visual Comparisons tested on Urban100~\cite{huang2015single-urban100} for HEVC codec~\cite{sze2014high-HEVC} at x4 scale of QP=[32, 37, 42]. }
	\label{fig: visual2}
\end{figure*}

\begin{figure*}[h]
	\centering 
	\includegraphics[width=1\linewidth]{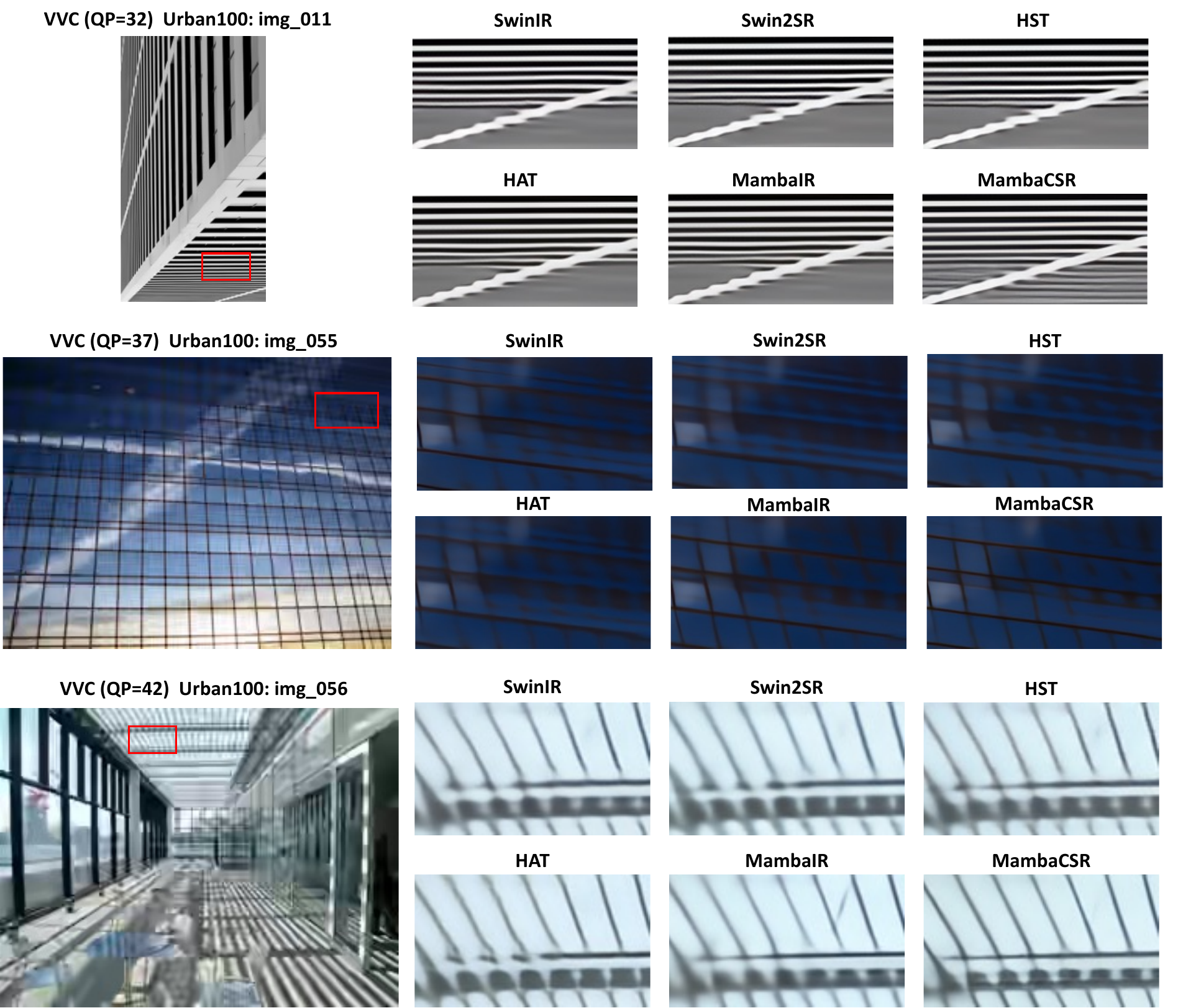}
	\caption{Visual Comparisons tested on Urban100~\cite{huang2015single-urban100} for VVC codec~\cite{bross2021overview-VVC} at x4 scale of QP=[32, 37, 42]. }
	\label{fig:visual_vvc}
\end{figure*}

In this section, we provide additional visual comparisons for the HEVC and VVC codec across three distortion levels: QP values from [32, 37, 42]. As depicted in Fig.~\ref{fig: visual2} and Fig.~\ref{fig:visual_vvc}, it is evident that our MambaCSR model demonstrates a significant advantage in mitigating compression artifacts, delivering a more structured restoration of textures and structural details compared to other approaches. In contrast, MambaIR falls short in preserving detailed local structural information, highlighting its limitations in capturing fine-grained features. These advancements can be attributed to our proposed hierarchical-interleaved scanning and position-aligned cross-scale scanning strategies, which enable more superior contextual information modeling than other existing models.

\end{document}